\documentclass[conference]{IEEEtran}
\IEEEoverridecommandlockouts
\usepackage{cite}
\usepackage{amsmath,amssymb,amsfonts}
\usepackage{algorithmic}
\usepackage{graphicx}
\usepackage{textcomp}
\usepackage{xcolor}
\def\BibTeX{{\rm B\kern-.05em{\sc i\kern-.025em b}\kern-.08em
    T\kern-.1667em\lower.7ex\hbox{E}\kern-.125emX}}
\begin{document}

\title{Exploring the potential of prototype-based soft-labels data distillation for imbalanced data classification}

\author{\IEEEauthorblockN{Radu-Andrei Rosu}
\IEEEauthorblockA{\textit{Faculty of Computer Science} \\
\textit{Alexandru Ioan Cuza University of Iasi}\\
\textit{Sentic Lab SRL}\\
Romania \\
}
\and
\IEEEauthorblockN{Mihaela-Elena Breaban}
\IEEEauthorblockA{\textit{Faculty of Computer Science} \\
\textit{Alexandru Ioan Cuza University of Iasi}\\
Romania \\
pmihaela@info.uaic.ro
}
\and
\and
\IEEEauthorblockN{Henri Luchian}
\IEEEauthorblockA{\textit{Faculty of Computer Science} \\
\textit{Alexandru Ioan Cuza University of Iasi}\\
Romania
}
}

\maketitle

\begin{abstract}
\emph{Dataset distillation} aims at synthesizing a dataset by a small number of artificially generated data items, which, when used as training data,  reproduce or approximate a machine learning (ML) model as if it were trained on the entire original dataset. Consequently, data distillation methods are usually tied to a specific ML algorithm. While recent literature deals mainly with distillation of large collections of images in the context of neural network models, tabular data distillation is much less represented and mainly focused on a theoretical perspective.  The current paper explores the potential of a simple distillation technique previously proposed in the context of Less-than-one shot learning. The main goal is to push further the performance of prototype-based soft-labels distillation in terms of classification accuracy, by integrating optimization steps in the distillation process. The analysis is performed on real-world data sets with  various degrees of imbalance. Experimental studies trace the capability of the method to distill the data, but also the opportunity to act as an augmentation method, i.e. to generate new data that is able to increase model accuracy when used in conjunction with - as opposed to instead of - the original data.
\end{abstract}


\section{Introduction}
Distillation has arisen as a research direction in the area of deep learning as a response to, on one hand, the high dimensionality of neural network structures and, on the other hand, the high dimensionality of the data needed to train such large structures. Consequently, two distinct research objectives are targeted: 
\begin{itemize}
    \item distilling the network, which aims at reducing the network size while achieving the same prediction accuracy; this is known as \emph{knowledge distillation}  \cite{hinton2015};
    \item distilling the data, which aims at reducing the size of the data needed to train the network while preserving the prediction accuracy under the same network structure; this is known as \emph{data distillation} \cite{dataDistillation2018}.
\end{itemize}

Knowledge distillation is strongly motivated by the need to reduce the computational burden at \textit{prediction} time - when the model is actually used in practical applications. Dataset distillation, which basically impacts the \textit{training} process, was firstly motivated by "a purely scientific question of how much data is encoded in a given training set and how compressible it is" \cite{dataDistillation2018}, with a modest concern for practicality.

The concept of data distillation was first introduced in 2018 in \cite{dataDistillation2018}, where neural networks are used to synthesize a data set consisting of a large number of images, using a technique based on gradient descent, so that each artificial image obtained contains information synthesized from several original images. Prior to the publication of \cite{dataDistillation2018}, only the concept of \textit{knowledge distillation} was studied, dealing with the transfer of the capabilities of a complex model into a simpler  one. \cite{dataDistillation2018} proposes rather a compression process that is concerned  with  the data itself, not the model.

 Initially illustrated in the context of algorithms using  neural networks, this concept has henceforth expanded.
Recent research has described ways of implementing data distillation that make use of  variations of other ML algorithms.
\cite{LOS, softPrototypeskNN} show that it is theoretically possible for a data set consisting of a very small number of artificial instances - in some cases,
even smaller than the number of classes -, to train a high-performance classifier based on an extension of the  k-NN algorithm.

\cite{softPrototypeskNN} mentioned above catalysed our interest in addressing three research directions into data distillation based on soft-labels prototypes:
\begin{itemize}
\item to what extent does the imbalance in data negatively impact the quality of distilled data and how can this be mitigated? 
\item is the distilled data efficient/usable only in the context of a specific algorithm, in this case the k-NN classifier?
\item what if the  original purpose of data distillation - minimizing the data set while \textit{preserving} the classification performance - is turned into a data augmentation technique - with an eye to  \textit{improving} the classification performance?
\end{itemize}

The rest of the paper is organized as follows. Section \ref{sec:relWork} describes related work in the general framework of data distillation, the "Less-Than-One -Shot learning" concept, eventually focusing on the k-Nearest Neighbor algorithm; works that aim at reducing the size of the training data that must be stored in the context of the "lazy learning" scenario are summarised. Section \ref{sec:contributions} describes the algorithm under   investigation and comes up with specific improvements. Section \ref{sec:experiments} presents the experiments we conducted in order to address the three research questions above, with one subsection dedicated to each of them.

\section{Related work}
\label{sec:relWork}
\subsection{Data distillation}
Initial experiments in data distillation \cite{dataDistillation2018} used neural networks to distill the MNIST data set consisting of 60000 images distributed in 10 classes, to only 10 synthetic images, one representative image for each class (digit), with a test-time recognition performance of 94\%, compared to 99\% for the original data set. It is noteworthy that these images have an artificial appearance, which diminishes the information perceived by the human eye about the numbers they represent, but at the same time, for a model based on a neural network, the amount of information that can be extracted from a single image of this kind is equivalent to the one extracted from several thousands instances of the initial data set.

\begin{figure*}[htbp]
\begin{center}
\includegraphics[width=1\textwidth]{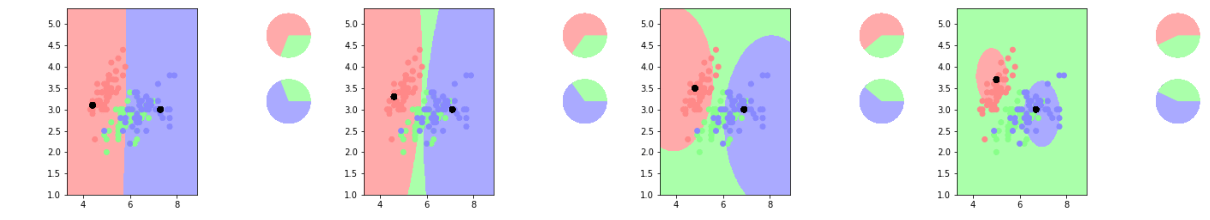}
\caption{kNN models fitted on 2 prototypes with soft labels. The pie charts represent the label distributions assigned to each of the 2 points.\cite{softLabelNeural}}
\label{fig:kNNbounderies}
\end{center}
\end{figure*}

A follow-up paper \cite{softLabelNeural} introduces the concept of \emph{soft-labels} as a way to reduce the number of instances in the distilled data set below the number of classes. Unlike hard labels, which always associate a training instance with one class and only one, soft labels allow to associate one training instance with several classes simultaneously, as a probability distribution. In this regard, \cite{softLabelNeural} proposes that, at the same time with the image distillation process, a label distillation process takes place.
The authors also introduce a new approach to data distillation by using the k-Nearest Neighbors (kNN) classifier. It is shown that the kNN algorithm can perform data classification that achieves high performance using only a small number of instances called \emph{prototypes}. These prototype instances can be obtained by selection, when they represent a subset of the entire dataset, or by generation, when they do not belong to the original dataset, but are artificially created. The latter is a better solution, as their features can be changed along the way while also receiving soft labels, in order to achieve the highest possible performance with the kNN classifier. 
Figure \ref{fig:kNNbounderies}, inserted for convenience from \cite{softLabelNeural}, illustrates the decision bounderies of a kNN model fitted on 2 distilled data items obtained using a combination of prototype generation and soft labels.

\subsection{Less-than-one -shot learning}
The \emph{soft label} concept was the cornerstone for the introduction of \emph{"LO" -shot learning} (\textit {Less Than One}) which makes possible for a model to learn to classify a number of N classes, using for training a data set consisting of only M instances, where M \textless N \cite{LOS}. Previously, there existed the concept of \textit {Few-shot learning}, ie supervised learning in which a model learns using a small data set for training, consisting of only a few training instances per class. The most extreme form of \textit {Few-shot} learning  was considered to be the so-called \textit {One-shot} learning, which used only one instance per class to train a model. Intuitively, the idea that a data set could be reduced more than that was considered impossible.

To demonstrate the feasibility of LO-shot learning, \cite{LOS} proposes a variation of the weighted kNN, called SLaPkNN (\textit {Soft-label Prototype k-Nearest Neighbors}). This variation of kNN calculates, for each instance to be classified, a soft label as a weighted sum of the  labels of the nearest k neighbors, where the weights are inverse proportional with the distance between the instance and the respective neighbor. As a result, the class predicted for the instance to be classified will be represented by the class with the highest value (or probability, if the values are scaled to fit a probability distribution). 
The authors demonstrate that 3 classes approximately collinear and equally spaced in a 2-dimensional space can be well separated by only 2 prototype instances. As a direct extension of this basic theorem, 
it can be demonstrated mathematically that a number of only 2 prototype instances can separate a large number of classes equally spaced from each other and positioned approximately along the line determined by those 2 prototypes. A generalization of this result will be used as basis for the algorithms presented in section \ref{sec:contributions}.

\subsection{Prototype selection/generation for nearest neighbours classification}
Prototype selection/generation in the context of kNN is not novel. KNN is known as a \textit {lazy} classifier, since it does not generate any model during the training stage, but stores all the data in the set, assigning them hard labels. This drives one big disadvantage of the algorithm, namely the need for a large storage space. In an attempt to improve this aspect, various studies have shown that it is possible for the kNN algorithm to perform well using only a small number of prototype instances, obtained by selection or generation. 

Regarding the selection of prototypes, \cite {i5} presents an extensive study on this topic and proposes a detailed taxonomy by grouping several selection methods. The main criterion for differentiating between methods, used in the taxonomy, is the type of selection that the method tries to make: removing data items from the decision boundaries that are most different from the majority in a class or retaining instances from the original data set that outline those boundaries while removing all internal instances within an area.


Also, a similar study exists in the case of prototype generation methods \cite{i6}, the main criterion for classifying the methods being the prototype generation mechanism. Thus, 4 types of methods are defined: centroid-based methods, methods based on the division of space, methods based on the re-labeling of classes, methods based on positioning adjustment.


None of these methods uses soft-labels. Unlike the method we exploit in this paper, one representative/prototype generated in the aforementioned studies captures information from a single class; thus, they are not able to summarise an entire data set by a number of instances less than the number of classes.

\section{Heuristic enhancements for soft-label prototype generation}
\label{sec:contributions}
\subsection{Soft-Label Prototypes}
Our work is based on soft-label prototype generation and the SLaPkNN algorithm, both presented in \cite {softPrototypeskNN} for LO-shot learning in the context of the kNN classifier.

The first step for prototype generation of the algorithm in \cite {softPrototypeskNN} is to find as few as possible distinct subsets of approximately collinear centroids in the original feature space, subsets which, together, form the set of all the centroids that are necessary to capture the original information in data. The purpose of this procedure is to obtain a number of lines, less in number than the total number of classes, so that any class belongs to approximately one of these lines. The subsets corresponding to all the centroids positioned on the same line will be therefore called prototypical lines. 

The second step of the algorithm is to generate soft labels for each line, so that the centroids representing the ends of a line contain distilled information about all the classes that the line represents. However, since the condition that classes must be positioned equidistantly along a line is difficult to fulfill in practice, for each prototype line, the problem of generating the two soft labels is designed in the form of a generic optimization problem, which does not take into account the distance between classes, but tries to maximize the influence of each class on its region along the line.

The method described above tends to perform well only if certain hypothesis on the data are satisfied: classes should be well delimited from each other and at equal distances. One of the main shortcomings for the prototype generation algorithm is that the lines depend only on the centroids of the classes and that their position cannot be changed along the way, but remain fixed after generation. Another major limitation of the algorithm is that, when generating the soft labels for a line, it is assumed that all classes must have regions of influence as uniform as possible along the line, regions given by a formula fixed to establish the boundaries between them. This hypothesis is usually violated in the case of imbalanced data sets.

Starting from these ideas, we aim in the following to optimize and further develop the components of the algorithm, so that the resulting final classifier performs well on as many real data sets as possible, including highly imbalanced data sets. The extensions are based on two ideas, corresponding to the two key elements of the algorithm, soft-label computation and prototype computation:
\begin{enumerate}
    \item iterative training of the initial prototype lines so that their soft labels improve along the way; we will call this procedure \emph{soft-label optimization};
    \item creating an ensemble of several sets of lines, with different weights, in the \textit {boosting} manner. Incorporating the boosting strategy for prototype generation encourages the generation of new prototypes in subspaces of the data set which are not well represented by the existing prototypes. This idea exceeds the context of LO-shot learning by generating a higher number of prototypes, with the aim of better covering the feature space in data and ultimately serve for data augmentation. This method also incorporates the optimization at step 1).  
\end{enumerate}

\subsection{Optimization of soft-labels}
The basic algorithm for generating soft labels for the prototypical lines always uses the same formula for establishing the regions of influence of classes along a line, based on the midpoints between two consecutive classes \cite {softPrototypeskNN}. As this leads to limited possibilities in positioning the decision boundaries between classes, a possible improvement of the algorithm in this regard is described below. 

In order to provide greater flexibility to the positioning of the class influence along a line, the boundaries between 2 consecutive regions will be represented by a point calculated by means of weights associated with the corresponding classes. Thus, the following formulas for computing the position of a boundary point based on class influence are considered, accompanied by a hypothetical graph of a line segment consisting of 3 points corresponding to some classes (Figure \ref{fig:P}):
\vspace{0.2cm}
\begin{figure}[htbp]
\centering
\includegraphics[width=5cm]{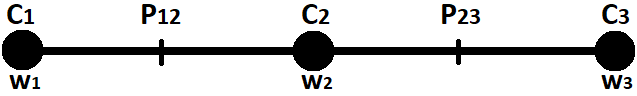}
\caption{Hypothetical graph of a line segment consisting of 3 centroids}
\label{fig:P}
\end{figure}

\begin{itemize}
    \item \(d(C_{1}, P_{1,2}) = d(C_{1}, C_{2}) \cdot \frac{w_{1}}{w_{1} + w_{2}}\) - where \(w_{1}\) represents the weight for class \(C_{1}\), and \(w_{2}\) represents the weight for class \(C_{2}\);
    \item \(d(C_{2}, P_{2,3}) = d(C_{2}, C_{3}) \cdot \frac{w_{2}}{w_{2} + w_{3}}\) - where \(w_{2}\) represents the weight for class \(C_{2}\), and \(w_{3}\) represents the weight for class \(C_{3}\);
    \item \(d(C_{i}, P_{i,i+1}) = d(C_{i}, C_{i+1}) \cdot \frac{w_{i}}{w_{i} + w_{i+1}}\) - the general form.
\end{itemize}

As for the values of the weights that determine the position of the borders of the regions, they will be gradually adjusted, through an iterative process, with a predefined number of steps. 

For each iteration dedicated to changing the influence of classes, a part of the training data is chosen to obtain predictions using the prototypical lines corresponding to that stage, and for each wrong prediction, but which at least belongs to the same line with the correct class, the following weight updates are performed, depending on a parameter \(\alpha\), which corresponds to the updating rate (a very small number, for example 0.01):
\begin{itemize}
    \item \(w_{g} = w_{g} - \alpha\) - where \(w_{g}\) represents the weight of the class wrongly assigned;
    \item \(w_{c} = w_{c} + \alpha\) - where \(w_{c}\) represents the weight of the correct class.
\end{itemize}

At the end of the weight update stage, the labels of the prototype lines containing classes whose weights have been modified are recomputed. Intuitively, the idea behind this iterative procedure is for the class regions to gradually change their size so that, at the end of the last iteration, the decision boundaries of the classifier are as accurate as possible. Although this method helps to alleviate the problem raised by the hypothesis of strict uniform regions present in the basic algorithm, the problem of dependency only on class centroids still remains valid. To improve the latter, a \textit{boosting} solution will be proposed in the next section of this paper.

\subsection{Prototypes generation using boosting}

The AdaBoost algorithm is a classic \textit{boosting} algorithm that can be used in combination with other classification algorithms in order to increase their performance. 
The main idea is to create, iteratively, a linear combination of classifiers called \textit{weak}, a combination that is itself a compound classifier, so that the rate of classification errors is gradually minimized. Originally designed only for binary classification problems, it was later extended so that it can be applied to any data set, regardless of its number of classes.

In this regard, \cite{i8} proposes a generalized variant of the AdaBoost algorithm, adapted for any number of classes. AdaBoost is an iterative algorithm, which uses classifiers that work with weighted instances during training. Initially, before the first iteration, it assigns equal weights to all the training instances, and then, at each iteration, these weights are adjusted according to the results of the weak classifier corresponding to that iteration. The basic idea is that the weights of the instances misclassified by the weak classifier from a certain iteration are increased, so that the classifier corresponding to the next iteration is more focused on classifying them correctly. Also, these weak classifiers corresponding to distinct iterations are applied with certain weights which dictate their contribution to the linear combination that represents the final classifier.

The weights corresponding to the weak classifiers are calculated according to the errors they obtain in the corresponding iterations. Depending on the weight of the classifier from a certain iteration, the new weights for the misclassified data instances are also calculated. At the end of the iteration, the weights of all instances are normalized. As regards the actual formulas, \cite{i8} proposes and rigorously justifies the following:

\begin{itemize}
    \item \(W_{cl}^{(i)} = \log\frac{1 - err^{(i)}}{err^{(i)}} + \log(N - 1)\) - where \(W_{cl}^{(i)}\) is the weight of the classifier at iteration \textit{i}, \(err^{(i)}\) is the weighted error at iteration \textit{i}, and \textit{N} is the number of classes in the data set;
    \item \(w_{in}^{(i+1)} = w_{in}^{(i)} \cdot \exp(W_{cl}^{(i)})\) - where \(w_{in}^{(i)}\) is the weigyt of a data instance wrongly classified at iteration  \textit{i}, and \(W_{cl}^{(i)}\) the weight of the classifier at iteration \textit{i}.
\end{itemize}

In order to be able to optimize, through AdaBoost, the algorithm for generating the prototype lines, a modification to the formula that computes the centroid is required. Thus, in the basic version of the algorithm, the coordinates of the centroid are always calculated as the arithmetic mean of the attributes of the corresponding instances. In order to introduce the concept of weights necessary at training, element fundamental within the AdaBoost algorithm, the arithmetic mean for calculating the centroid coordinates will be changed to a weighted arithmetic mean. Thus, all instances in the training data set will have weights, initially equal. At each AdaBoost iteration, the prototypical lines are generated using the iteration weights, and then the weighted error over the instances in the training set given by the HSLaPkNN classifier is calculated based on these lines. Based on this weighted error, the weight of the classifier corresponding to the iteration and the updates of the data weights are calculated in the classic AdaBoost manner. 

The final classifier will consist in a repeated application of the HSLaPkNN algorithm for each set of lines from each iteration. The result of this classifier will be given by the linear combination of all the results of the algorithm accompanied by the weights corresponding to the iterations.

Note that this \textit{boosting} approach involves keeping in memory all the sets of lines generated during iterations, accompanied by the appropriate weight for each one. Such an approach violates the objective of LO-shot learning, where the number of instances in the distilled data set is less than the total number of classes in the set. However, this method was specifically designed to generate a set of several prototypes that can, together, better represent the data distribution, meaningful for the classifier. The experimental results show that only 10 iterations of AdaBoost can help significantly increase the performance of the classifier, while still significantly reducing the size of the original data set.

\section{Experimental analysis}
\label{sec:experiments}
The experiments will be described in 3 distinct subsections distinguishing among the different research purposes. In each of these, the results of experiments performed on 10 sets of tabular data will be presented. These 10 data sets will remain constant throughout the experiments, so that, in the end, comparisons can be made between results obtained in different subsections. 
Six of the data sets aim at binary classification, while 4 sets have several classes. 

Binary data sets are all unbalanced, each with a different degree of imbalance. These sets range from a relatively moderate imbalance, in which 22.94\% of the samples belong to the minority class, to a very large imbalance, in which only 2.36\% of the samples belong to the minority class. 

Regarding the imbalance of the data sets with more than 2 classes, it is worth mentioning one of them in which the minority class is represented by a percentage of only 1.38\% of the total number of cases, a percentage that can be considered an extreme imbalance degree. In the tables and graphs illustrating the experimental results, the data sets will always be divided into two groups, according to the criterion of whether or not they are binary, and then, within each group, they will be ordered in ascending order of their imbalance degree (\textit {Imbalance Ratio} - IR). 

Table \ref{tab:datasets} presents the characteristics of the data sets used in the experimental analysis:

\linespread{1.2}
\begin{table}[htbp]
\centering
\begin{tabular}{|c|c|c|c|c|}
    \hline 
    Data set & attributes & classes & instances & IR \\
    \hline
    ecoli1 & 7 & 2 & 336 & 3.36 \\
    ecoli2 & 7 & 2 & 336 & 5.46 \\
    glass2 & 9 & 2 & 214 & 7.94 \\
    glass5 & 9 & 2 & 214 & 22.78 \\
    yeast4 & 8 & 2 & 1484 & 28.1 \\
    yeast6 & 8 & 2 & 1484 & 41.4 \\
    iris & 4 & 3 & 150 & 1 \\
    wine & 13 & 3 & 178 & 1.5 \\
    glass & 9 & 6 & 214 & 8.44 \\
    ecoli & 7 & 8 & 336 & 71.5 \\
    \hline
\end{tabular}
\caption{The data sets used in the experimental analysis}
\label{tab:datasets}
\end{table}
\linespread{1.5}

The first set of experiments is dedicated to evaluating the quality of distilled data sets (section \ref{subSec:distillationQuality}). 

The two enhancements we brought to the original data distillation algorithm are evaluated under the HSLaPkNN classifier against the basic algorithm from \cite{softPrototypeskNN}. Also, in order to further highlight the potential applicability of distilled sets in practice, a comparison will be made between the performance of the HSLaPkNN algorithm and the performance of some traditional linear classifiers. In addition, a series of three-dimensional visualizations, obtained by the principal component analysis (PCA), have the role of illustrating the way in which the algorithms work.

Sections \ref{subsec:NN} and \ref{subsec:augment} are dedicated to evaluating the relevance of the distilled data sets in combination with other classifiers and verifying their use as an augmentation technique in order to increase the quality of the original data set. In these experiments the HSLaPkNN classification algorithm is never used. Instead, a series of neural networks will be used, with one hidden layer and with two hidden layers, using \textit {cross entropy} as loss function. This choice is justified by the fact that, in the case of distilled data sets, a classifier that accommodates soft labels is needed.

For all the experiments the performance of the classifiers will be assessed both by accuracy and by F-score. In order to prevent the phenomenon of \textit {overfitting}, k-fold cross-validation methods will be used each time, where \(k = 5 \). Detailed information on how the actual partitioning of the data set is performed will be provided in the following, for each experiment.

\subsection{Evaluating the quality of the distilled data}
\label{subSec:distillationQuality}
The first experiment is to comparatively determine the benefits the enhancements we propose bring for data distillation. 

Besides experimenting with the distilled data, we also train linear classifiers on the original data as a way to evaluate the separability of the classes.
These classifiers are represented by the 1NN algorithm, \textit {Linear Regression} (LR) and \textit {Linear Discriminant Analysis} (LDA). The choice of the 1NN algorithm as a comparison standard is justified by the close link at implementation level with the HSLaPkNN algorithm, the latter being a direct extension of the classical kNN algorithm. However, from the point of view of the decision boundaries between the classes, this comparison is not necessarly fair, as in the case of the HSLaPkNN classifier the classes of the data set must be approximately linearly separable. This is a limitation that is not present in the case of 1NN. This is the reason why, we chose as competing algorithms only linear classifiers. Regarding the division of the data set into training data and test data, a standard 5-fold cross-validation methodology is used. Thus, each data set is divided into 5 folds in a stratified manner. Each fold will have, in turn, the role of test data and the remaining 4 folds will have the role of training data. In this way, in the end, all instances of the set will have an associated prediction, and based on these predictions the accuracy and F-score are calculated and reported.

For determining the performance of the HSLaPkNN classifier on the distilled data sets we experiment
\begin{enumerate}
    \item by using the original distillation method;
    \item by applying the first optimization presented in this paper, namely the optimization of soft labels; 500 training iterations are used, and class weights are initialized proportional to their number of instances;
    \item by using the boosting method for data distillation; 10 iterations of boosting are used, and within each boosting iteration the optimization from scenario 2 is incorporated with 100 iterations for soft-labels optimization.
\end{enumerate}
In all cases cross-validation is used. 

In case 1) above, 5 sets of distilled data will be formed, one by one, obtained from the union of 4 folds dedicated to training. Each of the 5 distilled sets will be used in turn to derive the predictions for the corresponding testing folds with the HSLaPkNN algorithm.

In cases 2) and 3), the distilled data sets will be generated on the union of only 3 folds out of a total of 4 dedicated to training. The fourth training fold will be used for soft-label updates in the optimization step to obtain predictions on data unseen at prototype generation. At the end, based on the 5th fold, strictly dedicated to testing, the predictions used to calculate quality metrics will be obtained. Within each combination of 4 folds for training, each fold will have, in turn, the role of fold dedicated to changing the weights of the classes during the iterative process. Thus, for each combination of 4 folds dedicated to training, 4 different distilled data sets will be obtained, and at the end 4 different predictions for each instance of the initial data set. Taking this into account, in the case of the iterative soft-label optimization, the values of accuracy and F-score will be calculated as an average of 4 values (one value for each set of predictions). The comparison with algorithm 1) is thus detrimental to cases 2) and 3) since prototypes are basically generated only on 3 folds and not 4; however, this complex experimental setting is necessary in cases 2) and 3) in order to avoid overfitting soft-labels to the training data.

\begin{figure}[htbp]
\centering\includegraphics[width=9cm]{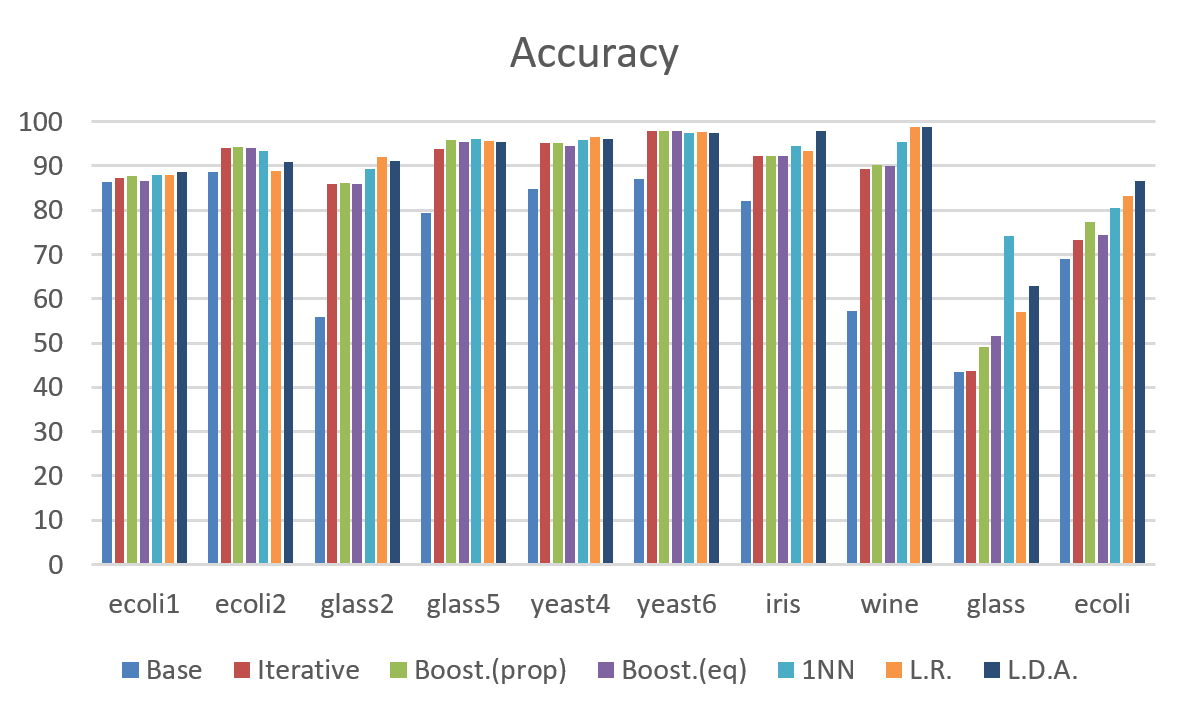}
\caption{Experimental results in terms of accuracy: Base = the algorithm proposed in \cite{softPrototypeskNN}, Iterative = iterative soft-label optimization, Boost = boosting optimization that produces 10 times more distilled instances; all distilled data sets are evaluated under the HSLaPkNN classifier. 1NN, LR and LDA are evaluated on the original data.}
\label{fig:resultsAcc}
\end{figure}

\begin{figure}[htbp]
\centering
\includegraphics[width=9cm]{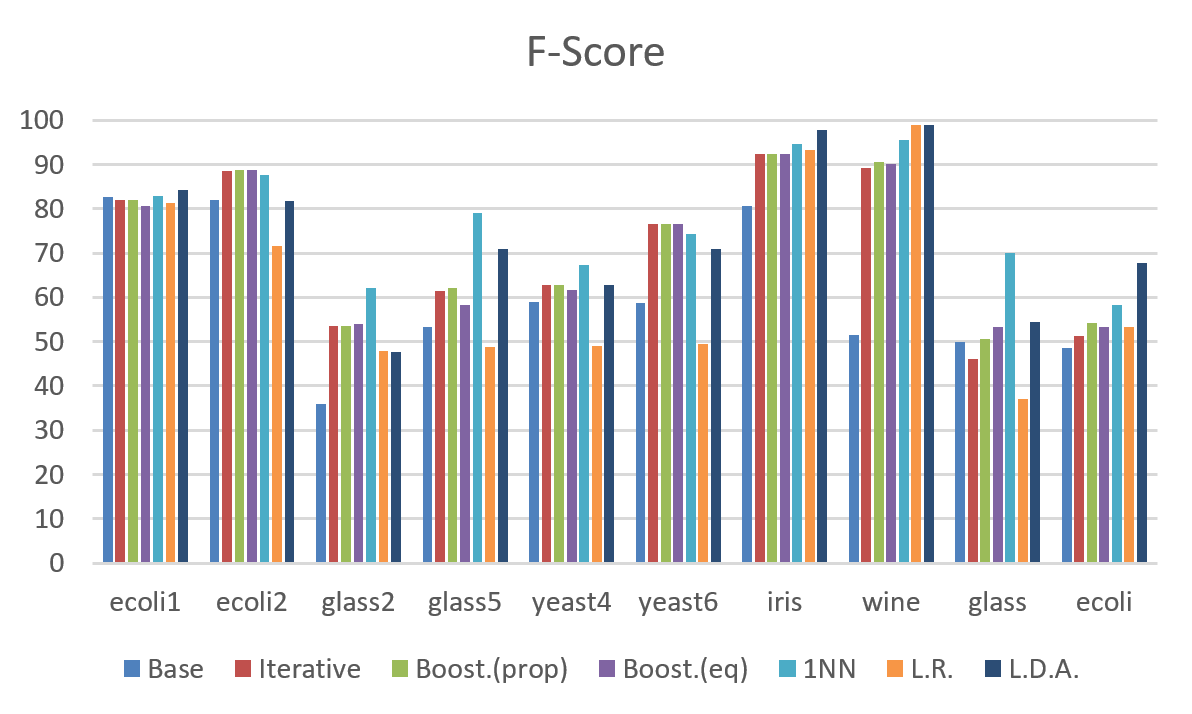}
\caption{Experimental results in terms of the F-score}
\label{fig:resultsF}
\end{figure}

All the experimental results are illustrated in Figures \ref{fig:resultsAcc} and \ref{fig:resultsF}.

For most distilled data sets, the optimization steps proposed in the paper bring a significant increase in quality metrics compared to the basic method. In this sense, the most significant are sets \textit{glass2}, \textit{yeast6} and \textit{wine}. In the case of the latter, there is an increase of more than 30\% both in accuracy and in the F-score. It is also worth noting that for a number of binary data sets, such as \textit{ecoli2} or \textit {yeast6}, the results obtained by the HSLaPkNN classifier are surprisingly even higher than those obtained by any of the three traditional classifiers, trained on the original data set.

In the case of all 6 binary data sets, the boosting optimization does not have much effect. Thus, in their case, high performance can be achieved only through the first optimization. This is a favorable aspect, because the size of the distilled dataset is kept very small. Thus, it can be concluded that the information of the binary imbalanced data sets can be synthesized by an artificial data set consisting of only two instances.

In addition, it should be noted that in the case of these binary data sets, the degree of imbalance is not necessarily a factor influencing the quality of the artificial sets. In this sense, the data sets \textit {yeast4} and \textit {yeast6} have the highest degree of imbalance and yet their corresponding distilled sets lead to considerably better performance than in the case of the distilled versions for \textit {glass2} or \textit {glass5}. This is probably due to the fact that \textit {yeast} sets have many more instances than \textit {glass} sets. Thus, distillation algorithms can generate artificial instances that accurately approximate the real decision boundaries.

On the other hand, in the case of data sets with more than two classes, although there is no example in which the performance of standard classification algorithms is fully exceeded, the results are still very good. If in the case of data sets \textit {iris} and \textit {wine} only the iterative optimization of soft labels is still sufficient to generate the highest quality distilled data sets, in the case of data sets \textit {glass} and \textit {ecoli}  the importance of boosting optimization is evident. This is most likely justified by the number of classes in the set and, implicitly, by the number of lines generated. In this sense, the first two sets mentioned have only 3 classes; thus, in their case, only one prototype line is generated, similar to binary data sets. Instead, the data sets \textit {glass} and \textit {ecoli} have 6 and 8 classes, respectively; in both cases, a set of 3 prototype lines will be generated. Since boosting optimization is based on moving the position of the set of lines in the feature space, it is understandable why a set consisting of a larger number of lines, with much more possibilities to move in space, will benefit more from this optimization.

Table \ref{tab:dims} illustrates the dimension of the original sets and of the distilled sets. The original version and the iterative soft-label optimization versions generate the same number of prototypes. Since the boosting algorithm is executed for 10 iterations, the number of prototypes generated in the case of boosting-enriched distilled data set is ten times higher for each dataset and is not illustrated in the table.

\linespread{1}
\begin{table}[htbp]
\centering
\begin{tabular}{|c|c|c|c|c|c|c|}
    \hline 
    Data set & \#instances & \#lines & \#prototypes &\%prototypes \\
    \hline
    ecoli1 & 336 & 1 & 2 & 0.59\% \\
    ecoli2 & 336 & 1 & 2 &  0.59\%\\
    glass2 & 214 & 1 & 2 & 0.93\% \\
    glass5 & 214 & 1 & 2 & 0.93\% \\
    yeast4 & 1484 & 1 & 2 & 0.13\% \\
    yeast6 & 1484 & 1 & 2 & 0.13\% \\
    iris & 150 & 1 & 2 & 1.33\% \\
    wine & 178 & 1 & 2 & 1.12\% \\
    glass & 214 & 3 & 6 &  2.80\%  \\
    ecoli & 336 & 3 & 6 &  1.78\% \\
    \hline
\end{tabular}
\caption{The original data set size versus the distilled data set size.}
\label{tab:dims}
\end{table}
\linespread{1.5}

\begin{figure}[htbp]
\centering
\includegraphics[width=9cm]{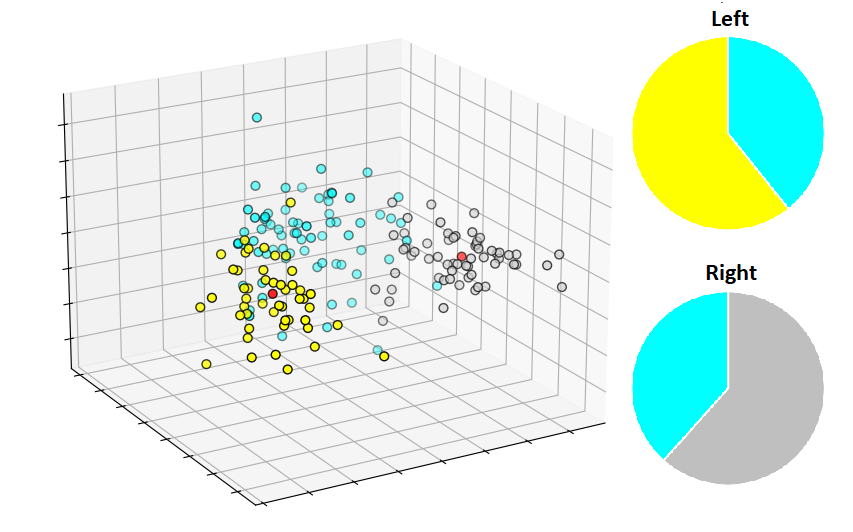}
\caption{PCA visualisation of the distilled prototypes (red) obtained with the iterative soft-label optimization, for the \textit{wine} data. }
\label{fig:PCA1}
\end{figure}

\begin{figure}[htbp]
\centering
\includegraphics[width=7cm]{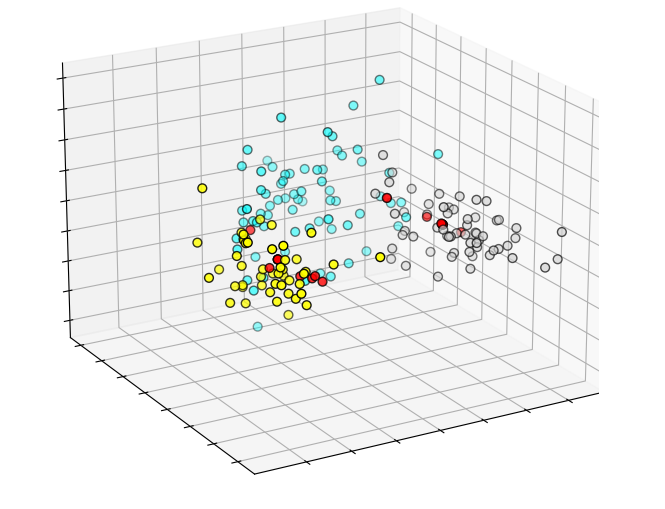}
\caption{PCA visualisation of the distilled prototypes (red) obtained with the boosting variant for the \textit{wine} data.}
\label{fig:PCA2}
\end{figure}

Figures \ref{fig:PCA1} and \ref{fig:PCA2} illustrate the distribution of the original data and the prototypes for the \textit{wine} data set.

\subsection{Evaluating the quality of the distilled data in the context of other classifiers}
\label{subsec:NN}

\begin{figure}[htbp]
\centering
\includegraphics[width=9cm]{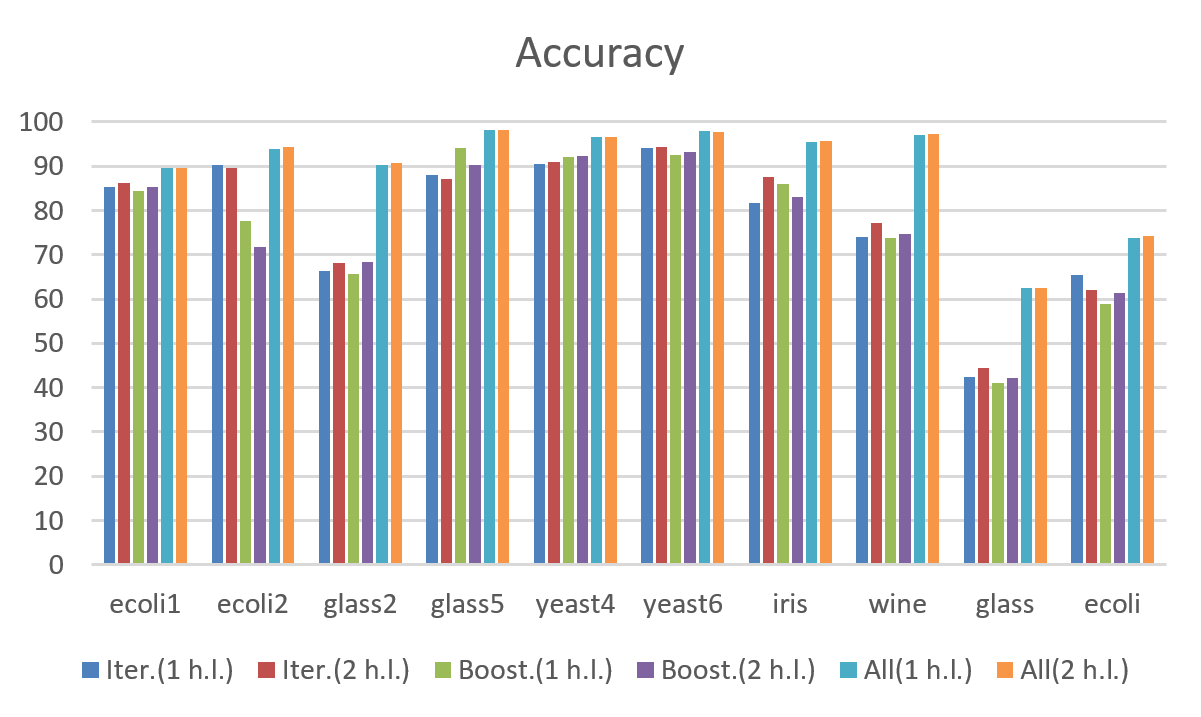}
\caption{The accuracy recorded with neural networks}
\label{fig:NNAcc}
\end{figure}

\begin{figure}[htbp]
\centering
\includegraphics[width=9cm]{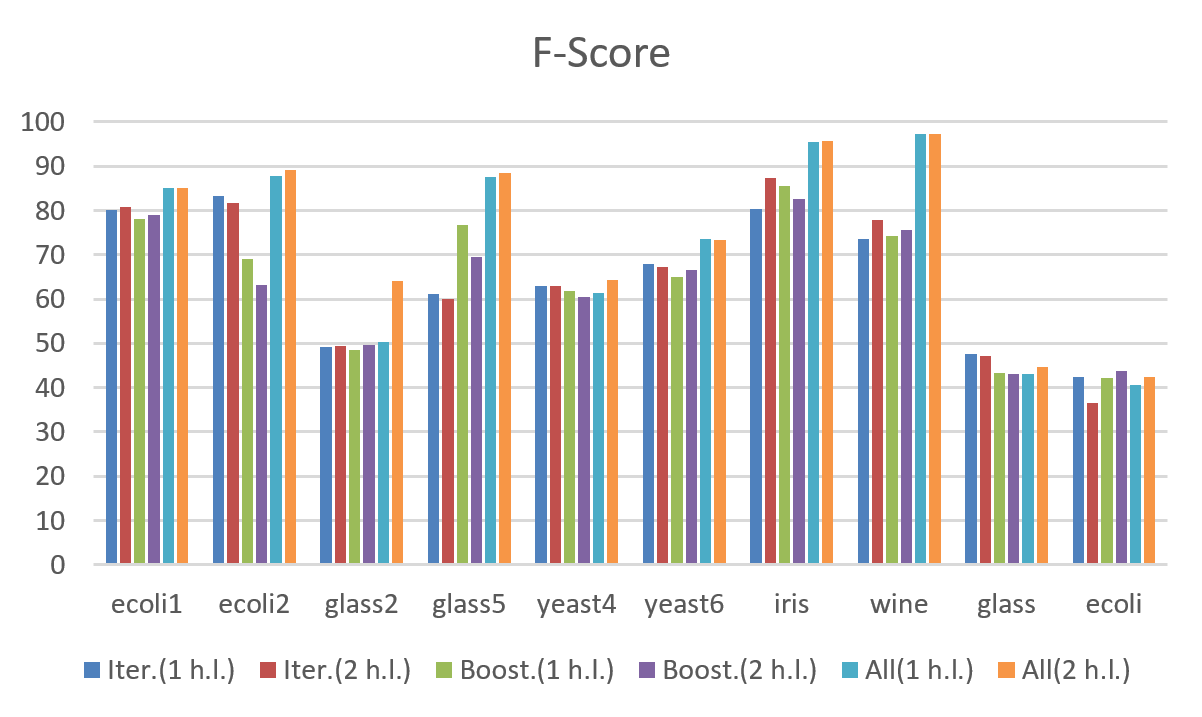}
\caption{The F-score recorded with neural networks}
\label{fig:NNF}
\end{figure}

Neural networks with 1 and 2 hidden layers were used to evaluate the quality of the distilled data. Figures \ref{fig:NNAcc} and \ref{fig:NNF} illustrate the comparative performance of the 2 networks when trained on:
\begin{itemize}
    \item the distilled set using the iterative optimization of soft-labels
    \item the distilled set using boosting
    \item the original/entire data set set.
\end{itemize}

The results show a satisfactory performance with only 2 to 6 (depending if binary or not the data set) distilled data items. Boosting does not seem to bring improvements, excepting 1 data set (glass 5).

\subsection{Evaluating the potential of the method for data augmentation}
\label{subsec:augment}
The same neural networks from the previous experiment were trained on the augmented data sets consisting of the original data and the distilled data that was generated using the boosting method. The same 5-fold cross-validation methodology was used where the distilled data is created only from the 4 folds used for training.

\begin{figure}[htbp]
\centering
\includegraphics[width=8cm]{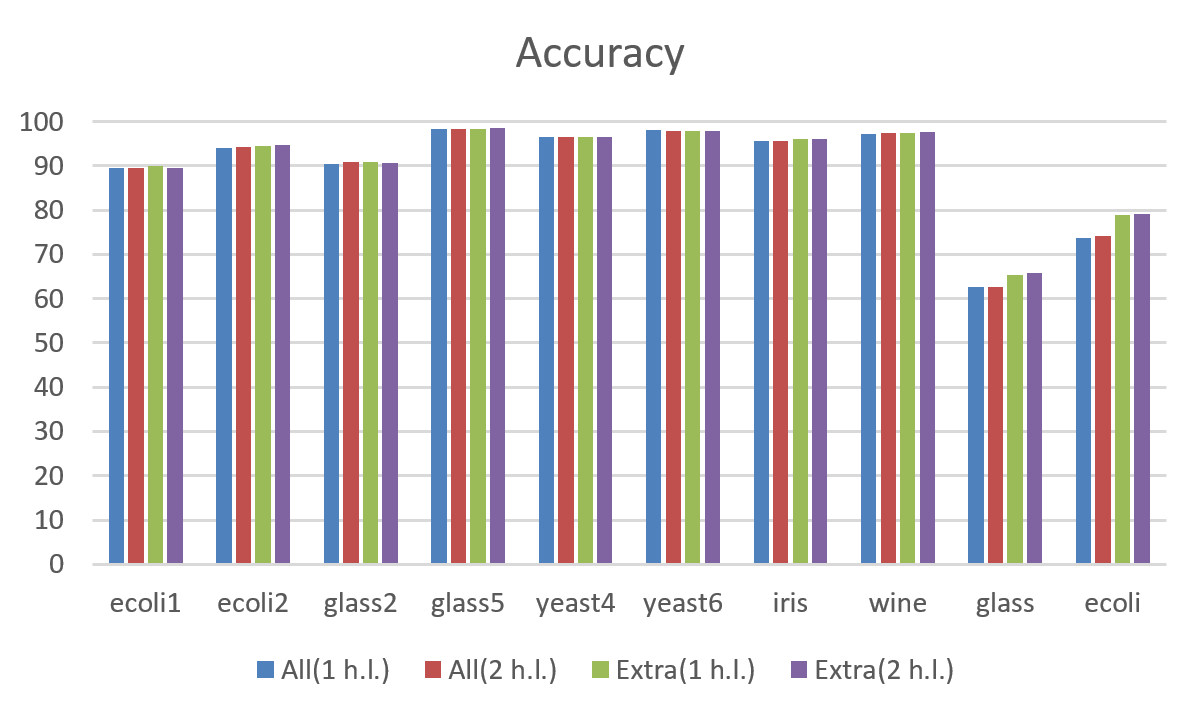}
\caption{Classification accuracy achieved with the neural networks on the original versus the augmented data sets}
\label{fig:augmentAcc}
\end{figure}

\begin{figure}[htbp]
\centering
\includegraphics[width=8cm]{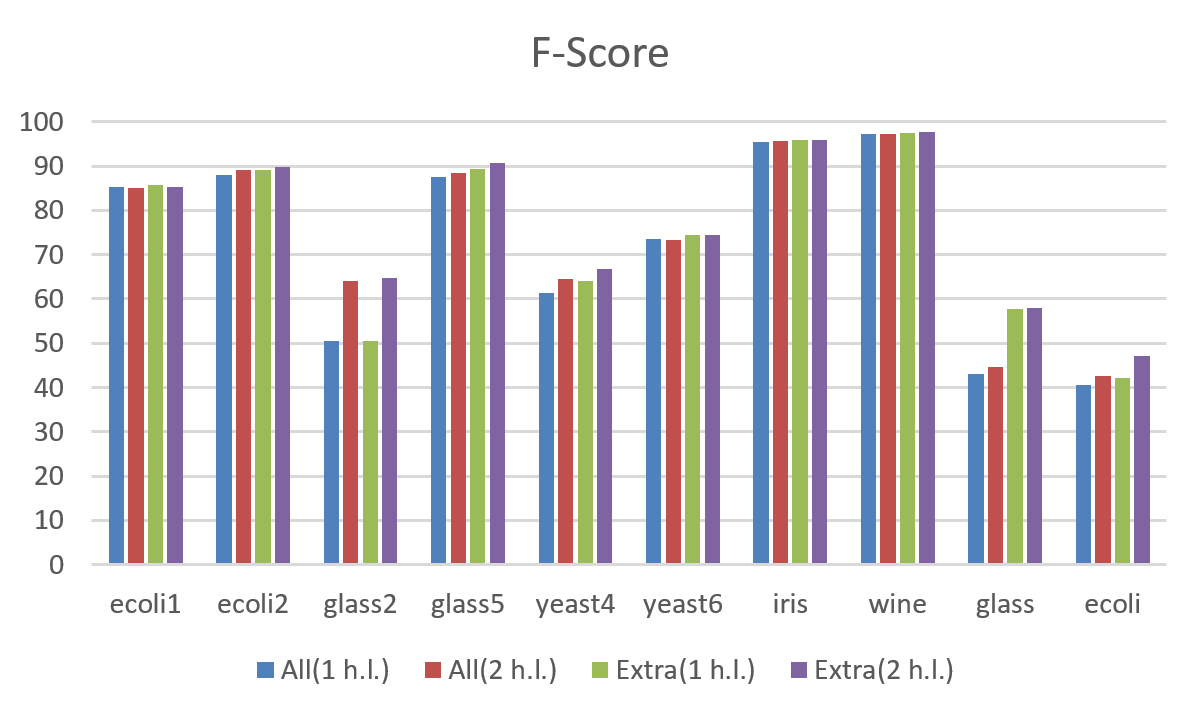}
\caption{The F-score achieved with the neural networks on the original versus the augmented data sets}
\label{fig:augmentF}
\end{figure}

The comparative results are presented in Figures \ref{fig:augmentAcc} and \ref{fig:augmentF} where the results obtained when using only the original data are shown first.

Analyzing Figures \ref{fig:augmentAcc} and \ref{fig:augmentF}, we can validate the hypothesis according to which the distilled soft-label instances obtained using the boosting procedure can be used as additional synthetic data to increase the quality of the classifier. Although the increase in performance is not so remarkable for all cases, there are still significant improvements in multi-class data sets such as \textit {ecoli} and \textit {glass}. This can be attributed to the fact that, in the case of these two data sets, many artificial instances are generated by applying boosting optimization. 
Thus, the data set \textit {ecoli} is expanded by approximately 18 \% of its original size, and the data set \textit {glass} is expanded by approximately 28 \% of its original size. Otherwise, no other set is extended by more than 14 \% of the base size.

\section{Conclusions}
The paper explores the data distillation potential in the context of classification of imbalanced data. In this context, an enhanced procedure for data distillation was proven to be able to distill the data set to a number of artificial soft-label instances at most equal to the number of classes (the Less-than-One shot scenario) with only small decrease in classification performance metrics recorded in the context of classification algorithms others than the one used in the process of distillation. A form of boosting was proposed to create more artificial instances that can represent better the distribution in the original data. As future direction we would like to extend the research on the use of the distillation algorithms in order to generate additional synthetic soft-label data, for augmentation purposes, to increase the quality of the classifiers. Thus, although we have shown in the experimental analysis that distilled sets through boosting optimization already have, to some extent, the ability to meet this goal, we believe that it may be possible to develop optimizations on the generative algorithm to improve more  the distilled data sets, in order to fulfill this direction.

\section*{Acknowledgment}
This paper is partially supported by the Competitiveness Operational Programme Romania under project number SMIS 124759 - RaaS-IS (Research as a Service Iasi).

\end{document}